# Global Stress Generation and Spatiotemporal Super-Resolution Physics-Informed Operator under Dynamic Loading for Two-Phase Random Materials

Tengfei Xing[1], Xiaodan Ren[1,2], Jie Li[1,2,*]

1. School of Civil Engineering, Tongji University, 1239 Siping Road, Shanghai 200092, China

2. The State Key Laboratory on Disaster Reduction in Civil Engineering, Tongji University, Shanghai 200092, China

*Corresponding author, E-mail: lijie@tongji.edu.cn

## Abstract

Material stress analysis is a critical aspect of material design and performance optimization. Under dynamic loading, the global stress evolution in materials exhibits complex spatiotemporal characteristics, especially in two-phase random materials (TRMs). Such kind of material failure is often associated with stress concentration, and the phase boundaries are key locations where stress concentration occurs. In practical engineering applications, the spatiotemporal resolution of acquired microstructural data and its dynamic stress evolution is often limited. This poses challenges for deep learning methods in generating high-resolution spatiotemporal stress fields, particularly for accurately capturing stress concentration regions. In this study, we propose a framework for global stress generation and spatiotemporal super-resolution in TRMs under dynamic loading. First, we introduce a diffusion model-based approach, named as Spatiotemporal Stress Diffusion (STS-diffusion), for generating global spatiotemporal stress data. This framework incorporates Space-Time U-Net (STU-net), and we systematically investigate the impact of different attention positions on model accuracy. Next, we develop a physics-informed network for spatiotemporal super-resolution, termed as Spatiotemporal Super-Resolution Physics-Informed Operator (ST-SRPINN). The proposed ST-SRPINN is an unsupervised learning method. The influence of data-driven and physics-informed loss function weights on model accuracy is explored in detail. Benefiting from physics-based constraints, ST-SRPINN requires only low-resolution stress field data during training and can upscale the spatiotemporal resolution of stress fields to arbitrary magnifications. Case studies demonstrate that our proposed framework achieves remarkable accuracy and generalization capability in both spatiotemporal stress field generation and super-resolution enhancement.

**Keywords**

Two-phase random materials, Global stress analysis, Spatiotemporal super-resolution, Stress concentration, Deep learning, Physics-informed

## 1. Introduction

The microstructure of two-phase random materials (TRMs) consists of two distinct materials that are randomly distributed in space, forming complex geometries and intricate interactions [1]. Such microstructures are commonly found in biological and biomimetic materials, including the air nanostructures in male Eastern bluebird feathers [2], synthetic bone materials that mimic human trabecular bone [3], and battery electrodes [4]. In practical applications, these materials are inevitably subjected to dynamic loading. Therefore, accurately and efficiently generating their dynamic responses is critical for the design of advanced composite materials.

The complexity and randomness of the microstructure in two-phase random materials (TRMs) pose significant computational challenges for stress analysis using conventional numerical approaches such as the finite element method (FEM). In recent years, most studies have focused on predicting the mechanical properties of materials under static loading. Researchers have improved classical convolutional architectures to capture the mapping relationship between the microstructure and stress field, while also employing transfer learning strategies to enhance cross-scale prediction capabilities [5]. However, these approaches still exhibit limitations in modeling the nonlinear response of materials. To overcome these constraints, researchers have successively introduced Transformer models [6] and graph neural networks (GNNs) [7]. The former leveraging sequential modeling paradigms to decode elastoplastic behaviors, while the latter employing topological inference frameworks to characterize complex failure mechanisms. Meanwhile, generative models have also provided powerful algorithmic tools for this field, such as the adversarial training-based microstructure-stress mapping methods [8] and stress generation frameworks incorporating diffusion processes [9]. However, existing methodologies exhibit a significant degree of homogeneity in problem settings, with research efforts predominantly focused on response prediction under static loading. A systematic solution for spatiotemporal analysis of mechanical properties under dynamic loading has yet to be established.

Existing approaches for dynamic response generation primarily focus on specific response components [10,11], while research on the global generation of spatiotemporal stress data remains limited. Recently, the emergence of video diffusion models has provided a novel technical framework to address this issue. Video diffusion models are built upon the diffusion process [12], generating high-quality videos through an iterative denoising procedure, thereby ensuring both temporal and spatial consistency [13]. Compared to methods such as TimeSformer [14] and StyleGAN-V [15], video diffusion models demonstrate significant advantages in video synthesis, editing, and generation tasks. Due to their capabilities in long-range dependency modeling, fine-detail preservation, and multimodal conditional control, these models have been widely applied in video generation [16], video editing [17], video super-resolution [18], and text-driven video synthesis [19]. The spatiotemporal stress data of materials under dynamic loading exhibit sequential characteristics analogous to video data, making video diffusion models a well-suited framework for this application. Consequently, integrating video diffusion models into stress field generation offers a promising research direction with potential applications in material science and mechanics.

Due to the small scale of the microstructure in two-phase random materials (TRMs), the spatiotemporal resolution of the acquired microstructural and dynamic stress evolution data is often limited in practical applications, which in turn constrains the accuracy of deep learning-based generation methods. In composite materials, failure is often accompanied by stress concentration [20,21]. When analyzing stress localization in microstructural regions, it is essential to obtain high-density stress data within these regions, necessitating an increase in the spatiotemporal resolution of stress fields. Existing deep learning-based video super-resolution (VSR) methods, such as TTVSR [22], iSeeBetter [23], and DiffVSR [24], primarily rely on supervised learning paradigms, where the resolution enhancement ratio is typically constrained by the correspondence between low- and high-resolution videos in the training dataset. In material dynamics problems, the key advantage of deep learning over FEM-based stress prediction lies in its computational efficiency. However, generating high-resolution stress datasets using FEM to facilitate the training of supervised models would drastically increase computational costs, thereby undermining the efficiency benefits of deep learning approaches. Moreover, unlike natural videos, spatiotemporal stress data are inherently governed by physical constraints [25]. The application of physics-informed neural networks (PINNs) in static elastic-plastic mechanics has matured in recent years [26,27], providing both theoretical justification and technical feasibility for employing PINNs in the spatiotemporal super-resolution of stress fields. Xing et al. first proposed the use of physics-informed operators for super-resolution stress analysis in random materials [28]. Subsequently, Oommen et al. extended this approach to the mechanical analysis of polycrystalline materials [29]. However, the super-resolution mechanical analysis of dynamic problems driven by physical information remains largely unexplored.

In this study, we propose a stress analysis framework for TRMs under dynamic loading. This framework can certainly be applied to stress analysis of materials with almost arbitrary microstructures. To incorporate both the microstructural characteristics of TRMs and the effects of dynamic loading, we introduce a tailored design for condition embedding. Then we develop a global spatiotemporal stress data generation method based on a diffusion model, referred to as STS-diffusion, which integrates the STU-net architecture. Furthermore, we analyze the impact of different attention positions on the generation performance of the model. Subsequently, we propose a spatiotemporal super-resolution physics-informed operator for stress data, termed ST-SRPINN, and investigate the effects of data-driven and physics-informed loss function weights on model accuracy. Notably, ST-SRPINN is an unsupervised approach, where the training process solely relies on the spatiotemporal stress data generated by STS-diffusion. By leveraging both data-driven and physics-informed constraints, the proposed method enables stress field resolution enhancement to an arbitrary magnification factor. Meanwhile, we examine the influence of loss function weight configuration and spatiotemporal resolution magnification on the error performance of ST-SRPINN.

The paper is structured as follows: Section 2 presents the global stress analysis framework for TRMs under dynamic loading in detail. Section 3 discusses the data results obtained from the proposed framework. Section 4 provides the conclusions.

## 2. Methodology

### 2.1. Generation method of spatiotemporal stress data based on diffusion model (STS-diffusion)

#### 2.1.1. Dataset generation

The microstructure of TRMs can be regarded as a probabilistic space, where any given sample represents a multivariate realization of a specific random field [30]. Consequently, the microstructure can be generated by segmenting a Gaussian random field with the same spectral density function as the original TRMs microstructure. In this study, a Gaussian stationary random field is generated using the stochastic harmonic function (SHF) method [31], which enables an efficient approximation of the two-point correlation function of the original microstructure with a limited number of summation terms, as detailed in **Appendix A**. Once the Gaussian stationary random field is obtained, horizontal segmentation is applied to generate microstructure images with the desired two-phase volume fraction. Considering the practical limitations in microstructure acquisition accuracy and the computational efficiency of STS-diffusion, the resolution of the generated TRMs microstructure images is set to 64 × 64, as illustrated in **Fig. 1**.

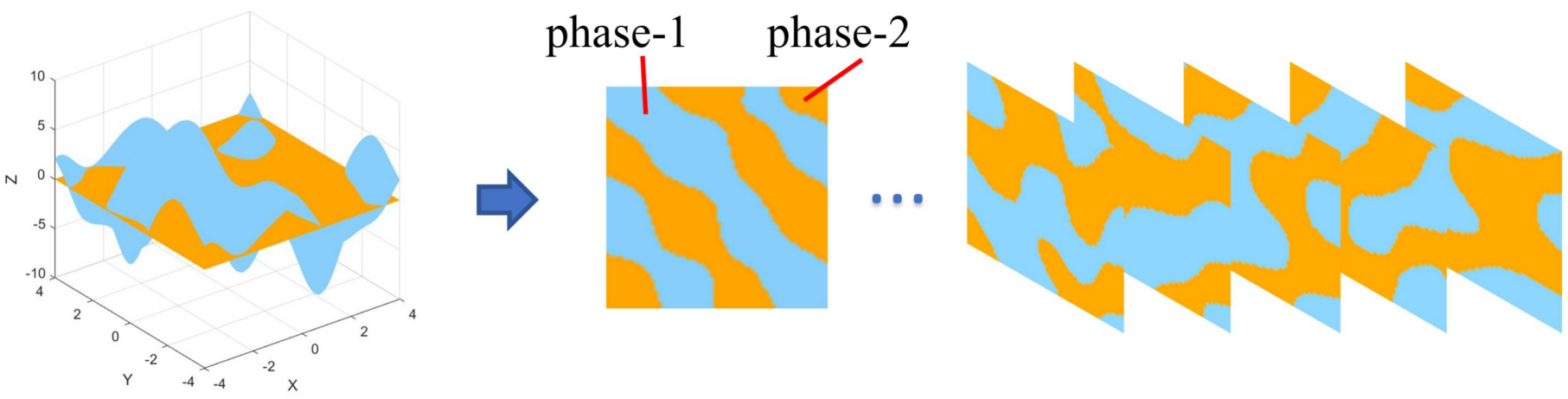

**Fig. 1.** Microstructure of TRMs.

In each denoising step of STS-diffusion, Gaussian noise generation is performed using the Space-Time U-Net (STU-net) architecture. Inspired by the Stable Diffusion framework proposed by Robin Rombach et al. [32], we concatenate the condition tensor directly with the original tensor along the channel dimension to construct the input tensor for STU-net, as illustrated in **Fig. 2c**.

The condition tensor $\boldsymbol{S}_{\mathrm{cond}}$ encapsulates both the microstructural conditions of TRMs $\boldsymbol{S}_{\mathrm{ms}}$ and the dynamic loading conditions $\boldsymbol{S}_{\mathrm{load}}$, and it can be expressed as:

$$\boldsymbol{S}_{\mathrm{cond}} = \boldsymbol{S}_{\mathrm{ms}} \oplus \boldsymbol{S}_{\mathrm{load}} \tag{1}$$

where $\oplus$ denotes the concatenation operator along the channel dimension.

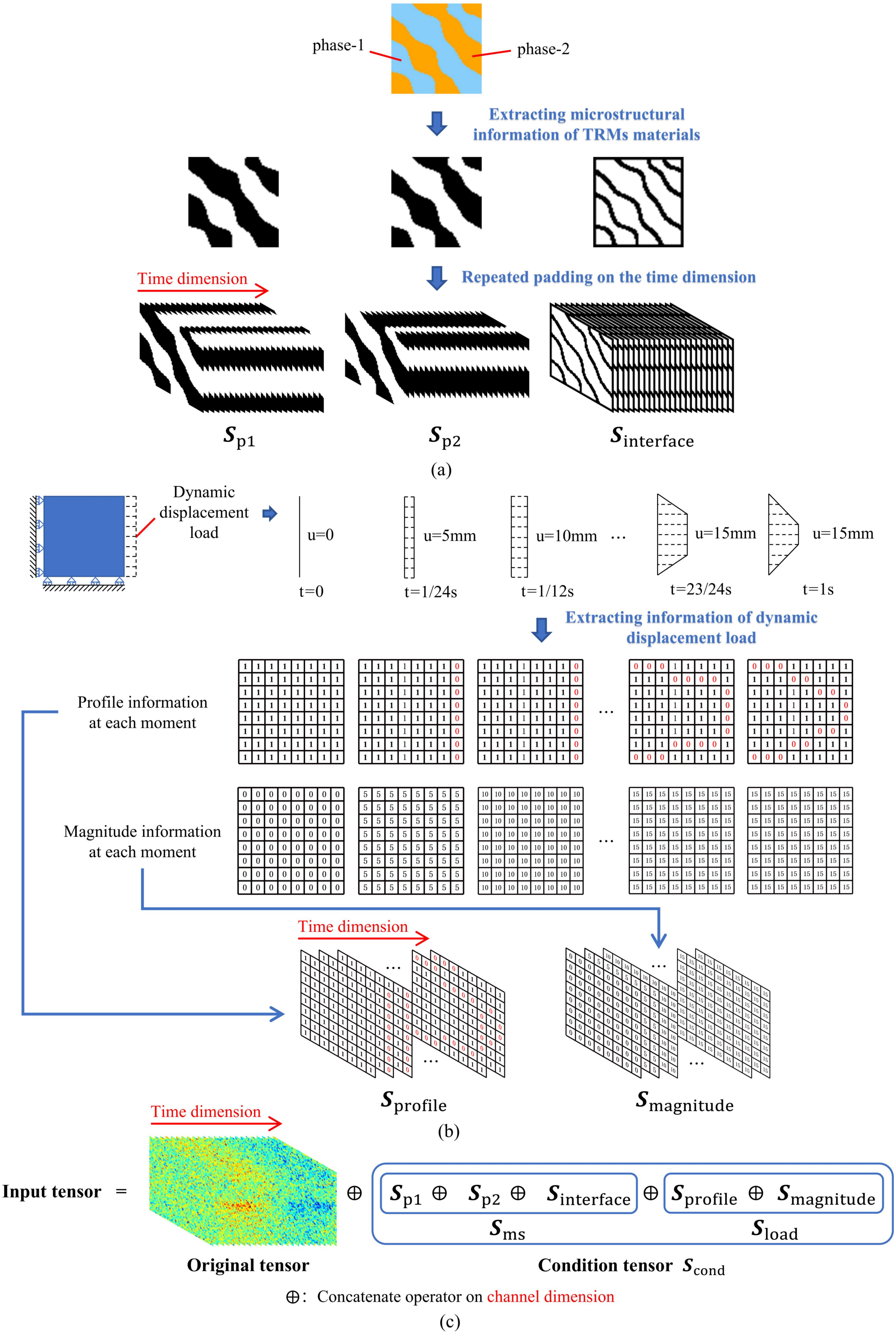

**Fig. 2.** Feature construction of input tensor.

Considering that interface effects can lead to stress concentration at the phase interfaces of TRMs, thereby promoting crack propagation along the interfaces and degrading material performance [33-35], $\boldsymbol{S}_{\mathrm{ms}}$ not only encodes the spatial distributions of different material phases ( $\boldsymbol{S}_{\mathrm{p1}}$ and $\boldsymbol{S}_{\mathrm{p2}}$ ) but also incorporates the location information of phase interface regions $\boldsymbol{S}_{\mathrm{interface}}$ . The location information is represented by tensors filled with Boolean values: a value of 1 (black) indicates that the corresponding position belongs to the specified region, whereas a value of 0 (white) denotes otherwise. The time dimension of $\boldsymbol{S}_{\mathrm{p1}}$ , $\boldsymbol{S}_{\mathrm{p2}}$ and $\boldsymbol{S}_{\mathrm{interface}}$ , structured as [batch (B), channel (C), time (T), height (H), width (W)], is populated by repeating the lower-dimensional features [B, C, H, W] across the temporal axis, as illustrated in **Fig. 2a**. $\boldsymbol{S}_{\mathrm{ms}}$ can be expressed as:

$$\boldsymbol{S}_{\mathrm{ms}} = \boldsymbol{S}_{\mathrm{p1}} \oplus \boldsymbol{S}_{\mathrm{p2}} \oplus \boldsymbol{S}_{\mathrm{interface}} \tag{2}$$

TRMs are subjected to loading through boundary displacements. $\boldsymbol{S}_{\mathrm{load}}$ includes both the normalized displacement profile information $\boldsymbol{S}_{\mathrm{load}}$ and the magnitude information of the applied load $\boldsymbol{S}_{\mathrm{magnitude}}$ . To obtain the normalized displacement profile, the displacement load at each time step is divided by the corresponding maximum displacement value at that time step. The normalized displacement profile across different time steps is represented as a Boolean tensor, where a value of 0 indicates the presence of the normalized displacement profile. Meanwhile, the load magnitude at different time steps is encoded as a constant tensor filled with the corresponding load magnitude values, as illustrated in **Fig. 2b**. $\boldsymbol{S}_{\mathrm{load}}$ can be expressed as:

$$\boldsymbol{S}_{\mathrm{load}} = \boldsymbol{S}_{\mathrm{profile}} \oplus \boldsymbol{S}_{\mathrm{magnitude}} \tag{3}$$

To enhance the generation accuracy of STS-diffusion, spatiotemporal stress data normalization is required. We designed an auxiliary network to restore the generated data from STS-diffusion. This auxiliary network predicts the maximum and minimum values ( $s_{\mathrm{max}}$ and $s_{\mathrm{min}}$ ) of the original spatiotemporal stress data by $\boldsymbol{S}_{\mathrm{cond}}$ , as illustrated in **Fig. 3**. The auxiliary network is a spatiotemporal feature extraction model that integrates 3D convolution with a multi-head self-attention mechanism to process input tensors of shape [B, C, T, H, W]. First, two layers of 3D convolution with ReLU activation and max pooling are applied to extract local spatiotemporal features while progressively reducing the data dimensionality. Next, the extracted convolutional features are flattened into a sequential representation, and a multi-head self-attention mechanism is introduced to capture long-range dependencies through global feature modeling. Finally, global average pooling is used to further reduce dimensionality, followed by two linear layers that map the features to the final two-scalar output per batch, with a dimension of [B, 2].

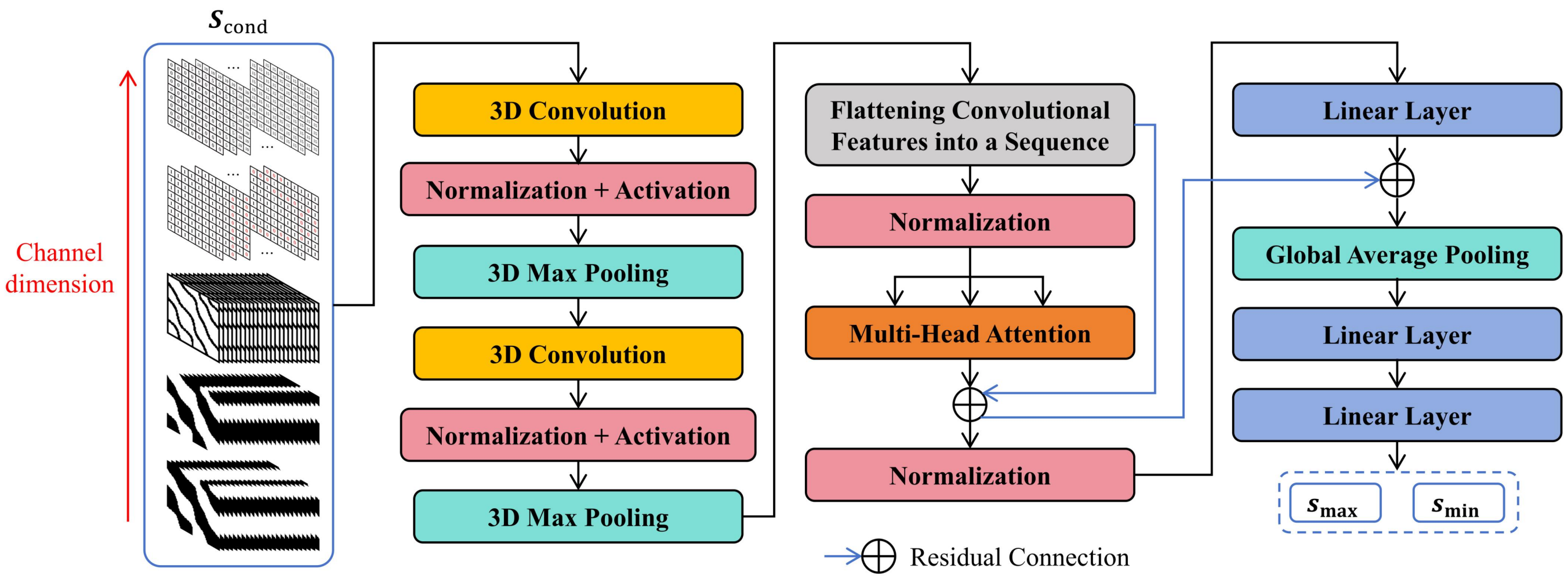

**Fig. 3.** Auxiliary network architecture.

### 2.1.2. Architecture

STS-diffusion primarily comprises a forward diffusion process and a reverse denoising process, as shown in **Fig. 4**. The forward diffusion process follows a Markov chain, in which Gaussian noise is incrementally added at each step [36]

$$q\left(\boldsymbol{s}_{t_d} \middle| \boldsymbol{s}_{t_d-1}\right) = \mathcal{N}\left(\boldsymbol{s}_{t_d}; \sqrt{1-\beta_{t_d}}\,\boldsymbol{s}_{t_d-1}, \beta_{t_d}\mathbf{I}\right), \quad q\left(\boldsymbol{s}_{1:T_d} \middle| \boldsymbol{s}_0\right) = \prod_{t_d=1}^{T_d} q\left(\boldsymbol{s}_{t_d} \middle| \boldsymbol{s}_{t_d-1}\right) \tag{4}$$

where $\beta_{t_d} \in (0,1)$ is a hyperparameter that controls the noise addition step size, while $t_d$ and $T_d$ represent the current and maximum time steps of STS-diffusion, respectively. The reverse denoising process also follows a Markov chain, where the model progressively denoises a randomly sampled Gaussian noise step by step until the final spatiotemporal stress field is reconstructed:

$$p\left(\hat{\boldsymbol{s}}_{T_d}\right) = \mathcal{N}\left(\hat{\boldsymbol{s}}_{T_d}; 0, \mathbf{I}\right), \quad p_{\boldsymbol{\theta}}\left(\hat{\boldsymbol{s}}_{t_d-1} \middle| \hat{\boldsymbol{s}}_{t_d}\right) = \mathcal{N}\left(\hat{\boldsymbol{s}}_{t_d-1}; \boldsymbol{\mu}_{\boldsymbol{\theta}}\left(\hat{\boldsymbol{s}}_{t_d}, t_d\right), \sigma_{t_d}^2\mathbf{I}\right), \quad p_{\boldsymbol{\theta}}\left(\hat{\boldsymbol{s}}_{0:T_d}\right) = p\left(\hat{\boldsymbol{s}}_{T_d}\right)\prod_{t_d=1}^{T_d} p_{\boldsymbol{\theta}}\left(\hat{\boldsymbol{s}}_{t_d-1} \middle| \hat{\boldsymbol{s}}_{t_d}\right) \tag{5}$$

where $\hat{\cdot}$ represents the tensor generated during the reverse denoising process and $\boldsymbol{\theta}$ represents the parameters of the neural network, while $\hat{\boldsymbol{s}}_{t_d}$ can be regarded as the original tensor input to the STU-net, as illustrated in **Fig. 2c**. During training, the normalization process of $\boldsymbol{s}_0$ can be expressed as:

$$\boldsymbol{s}_0 = \boldsymbol{s}_{\mathrm{FEM}} / \max\left\{\mathrm{abs}\left(\boldsymbol{s}_{\max}\right), \mathrm{abs}\left(\boldsymbol{s}_{\min}\right)\right\} \tag{6}$$

where $\boldsymbol{s}_{\mathrm{FEM}}$ represents the spatiotemporal stress data obtained from FEM calculations, while $\mathrm{abs}(\cdot)$

denotes the absolute value operator. $s_{\max}$ and $s_{\min}$ correspond to the maximum and minimum values within $s_{\text{FEM}}$, respectively. During the generation phase, the final generated data $s_{\text{result}}$ can be expressed as:

$$s_{\text{result}} = \hat{s}_0 \cdot \max\left\{\text{abs}\left(s_{\max}\right),\ \text{abs}\left(s_{\min}\right)\right\} \tag{7}$$

where $\hat{s}_0$ represents the generated data obtained from the reverse denoising process, and at this stage, $s_{\max}$ and $s_{\min}$ are obtained through the Auxiliary network.

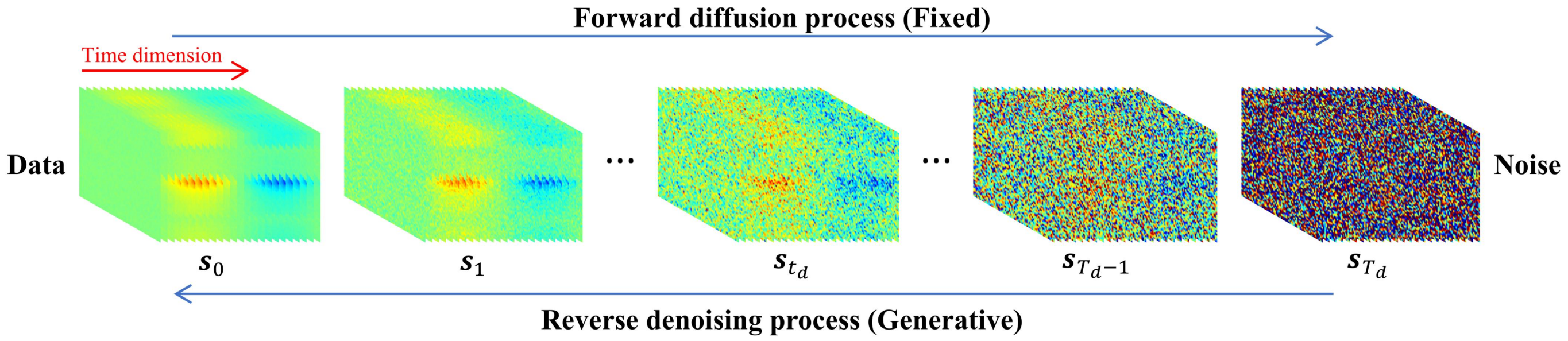

**Fig. 4.** The forward diffusion process and the reverse denoising process of diffusion model.

In this study, a STU-net with attention mechanism and temporal encoding was utilized as the denoising model, as shown in **Fig. 5**. The STU-net takes a five-dimensional tensor of size [B, C, T, H, W] as input. The network consists of an encoder (downsampling path), bottleneck, and decoder (upsampling path), and incorporates multi-head attention to enhance its modeling capability. The encoder progressively reduces the spatial resolution while increasing the number of channels to extract multi-scale spatiotemporal features. The bottleneck further improves feature representation by using ResNet block and a global attention mechanism. The decoder reconstructs the spatial resolution by fusing low-level details from the encoder using skip connections. Finally, after passing through a 3D convolution layer, the number of channel dimension is mapped to 1. Each level of the encoder and decoder can optionally incorporate spatial attention and temporal attention to enhance the ability of model to capture global spatiotemporal features. Specifically, temporal attention is optimized using rotary embedding to more effectively capture long-range temporal dependencies.

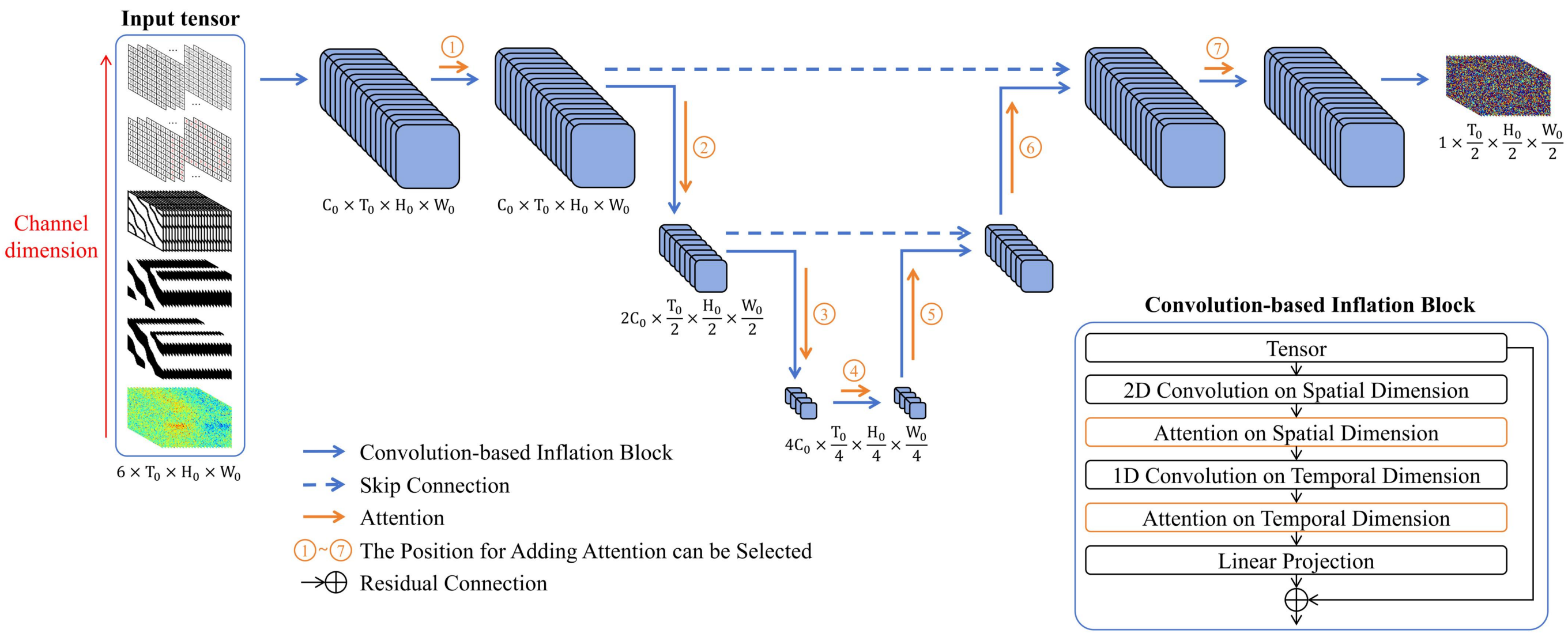

**Fig. 5.** STU-net architecture.

It is worth noting that considering the training efficiency of STS-diffusion, this study discusses the impact of the attention positions on training time and accuracy. The indexes (①, ②, ..., ⑦) in Fig. 5 indicate the positions where spatial attention and temporal attention can be selected. Algorithm 1 and Algorithm 2 illustrate the training and generation processes of STS-diffusion, respectively. $\varsigma$ and $\varsigma_\theta$ denote the noise sampled in the forward process and the noise generated by STU-net in the reverse denoising process. Define $\alpha_{t_d} = 1 - \beta_{t_d}$ and $\bar{\alpha}_{t_d} = \prod_{i=1}^{t_d} \alpha_i$.

---

**Algorithm 1: Training**

---

1: **repeat**

2: $\quad \boldsymbol{s}_0, \boldsymbol{S}_{\text{cond}} \sim q(\boldsymbol{s}_0, \boldsymbol{S}_{\text{cond}})$

3: $\quad t_d \sim \text{Uniform}(\{1, \ldots, T_d\})$

4: $\quad \varsigma \sim N(0, \mathbf{I})$

5: $\quad$ Take gradient descent step on $\nabla_\theta \left\| \varsigma - \varsigma_\theta \left( \sqrt{\bar{\alpha}_{t_d}} \boldsymbol{s}_0 + \sqrt{1-\bar{\alpha}_{t_d}} \varsigma, \boldsymbol{S}_{\text{cond}}, t_d \right) \right\|^2$

6: **until** converged

---

**Algorithm 2: Generation**

---

1: $\hat{\boldsymbol{s}}_{T_d} \sim N(0, \mathbf{I})$

2: **for** $t_d = T_d, \ldots, 1$ **do**

3: $z \sim N(0,\mathbf{I})$ if $t_d > 1$, else $z = 0$

4: $\hat{s}_{t_d-1} = \mu_\theta\left(\hat{s}_{t_d}, S_{\text{cond}}, t_d\right) + \sigma_{t_d} z = \frac{1}{\sqrt{\alpha_{t_d}}}\hat{s}_{t_d} - \frac{1-\alpha_{t_d}}{\sqrt{1-\bar{\alpha}_{t_d}}}\varsigma_\theta\left(\hat{s}_{t_d}, S_{\text{cond}}, t_d\right) + \sqrt{\frac{\left(1-\alpha_{t_d}\right)\left(1-\bar{\alpha}_{t_d-1}\right)}{1-\bar{\alpha}_{t_d}}} z$

5: **end for**

6: **return** $\hat{s}_0$

**2.2. Spatiotemporal super-resolution physics-informed operator of stress data (ST-SRPINN)**

In this study, a ST-SRPINN architecture was designed using elastic materials as an example, as shown in **Fig. 6**. The architecture consists of five parallel FFNs, each responsible for predicting displacement ($u_x$, $u_y$) and stress ($\sigma_{xx}$, $\sigma_{yy}$, $\sigma_{xy}$). Given the geometric equation in **Eq. (8)**, displacements and strains are equivalent outputs. Displacement is chosen here because it simplifies boundary condition constraints and reduces the number of network parameters compared to directly outputting strain components. The loss function integrates four components: observation point constraints (the stress values obtained through video diffusion model), displacement boundary conditions, the equilibrium equation **Eq. (9)**, and the constitutive relation **Eq. (10)**.

$$\boldsymbol{\varepsilon} = \frac{1}{2}\left(\nabla \otimes \boldsymbol{u} + \boldsymbol{u} \otimes \nabla\right) \tag{8}$$

$$\nabla \cdot \boldsymbol{\sigma} = \rho \frac{\partial^2 \boldsymbol{u}}{\partial t^2} \tag{9}$$

$$\boldsymbol{\sigma} - \lambda \mathrm{tr}\left(\boldsymbol{\varepsilon}\right)\mathbf{I} - 2\mu\boldsymbol{\varepsilon} = 0 \tag{10}$$

where $\nabla$ is the Hamiltonian operator, $\boldsymbol{u}$ is the displacement vector, $\boldsymbol{\varepsilon}$ and $\boldsymbol{\sigma}$ are the Cauchy strain and stress, respectively, $\mathrm{tr}(\cdot)$ is the trace of a matrix, $\mathbf{I}$ is the second-order identity tensor; $t$ is the time, $\rho$ is the material density. $\lambda$ and $\mu$ are Lamé coefficient, related to the Young's modulus $E$ and Poisson's ratio $\nu$:

$$\lambda = \frac{E\nu}{(1+\nu)(1-2\nu)}, \quad \mu = \frac{E}{2(1+\nu)} \tag{11}$$

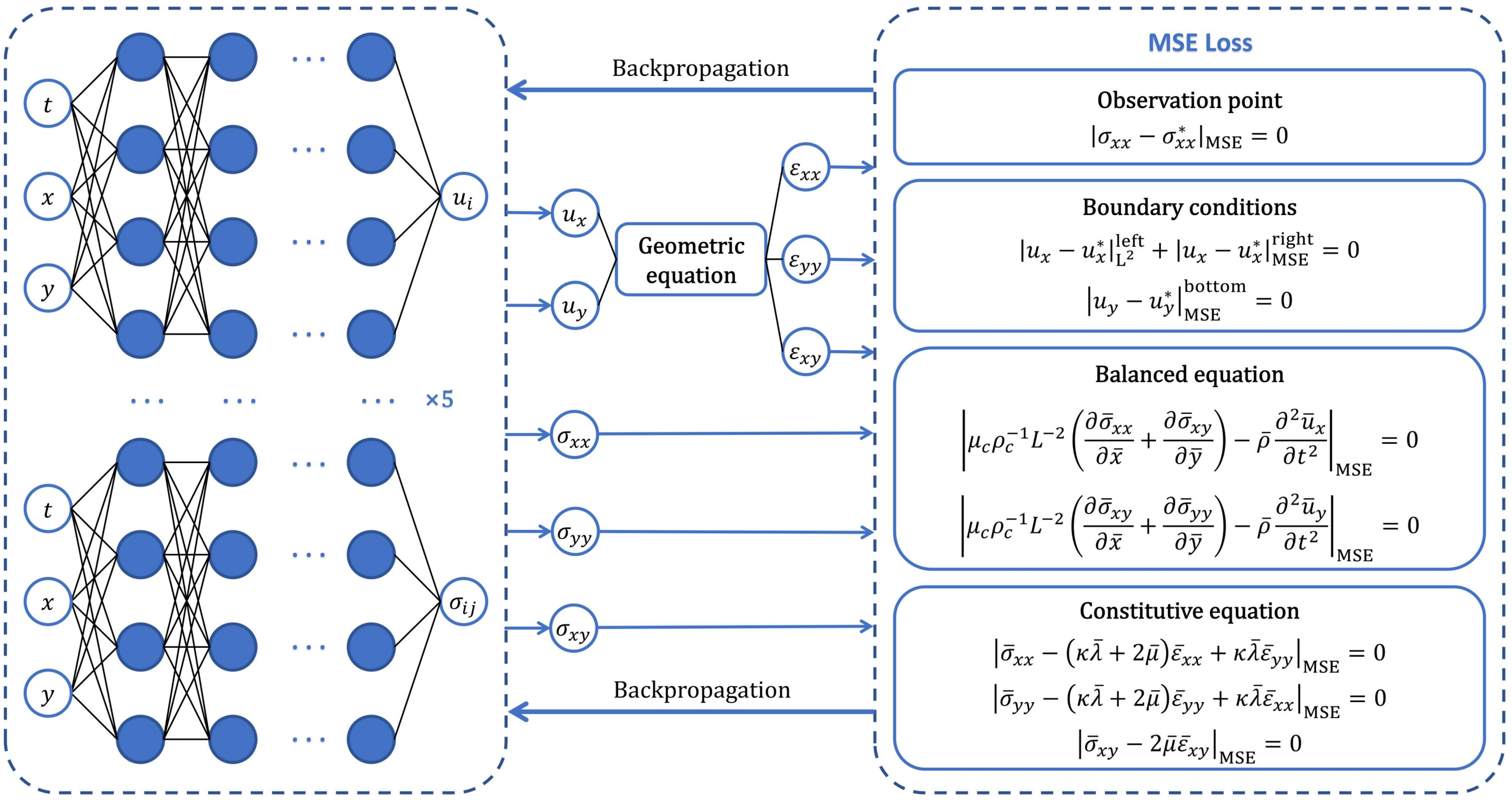

**Fig. 6.** PINN architecture.

To avoid multi-scale loss caused by significant differences in the orders of magnitude among parameters, it is necessary to normalize the parameters by converting them into a dimensionless form [37]

$$\bar{x}=\frac{x}{L},\quad \bar{y}=\frac{y}{L},\quad \bar{\rho}=\frac{\rho}{\rho_c},\quad \bar{\lambda}=\frac{\lambda}{\lambda_c},\quad \bar{\mu}=\frac{\mu}{\mu_c},\quad \bar{\boldsymbol{u}}=\frac{\boldsymbol{u}}{u_c},\quad \bar{\boldsymbol{\sigma}}=\frac{\boldsymbol{\sigma}}{\sigma_c} \tag{12}$$

where $L$ is the maximum value of the spatial coordinate interval; $\rho_c$, $\lambda_c$ and $\mu_c$ are the maximum values of $\rho$, $\lambda$ and $\mu$ within the spatial range, respectively:

$$\rho_c = \max_{x,y} \rho,\quad \lambda_c = \max_{x,y} \lambda,\quad \mu_c = \max_{x,y} \mu \tag{13}$$

To calculate $\sigma_c$ and $u_c$, first substitute **Eq. (8)** into **Eq. (10)**, yielding:

$$\boldsymbol{\sigma} = \lambda \nabla \cdot \boldsymbol{u} \mathbf{I} + \mu\left(\nabla \otimes \boldsymbol{u} + \boldsymbol{u} \otimes \nabla\right) \tag{14}$$

Convert the parameters in **Eq. (14)** into their dimensionless form:

$$\sigma_c \bar{\boldsymbol{\sigma}} = \lambda_c \bar{\lambda} u_c L^{-1} \bar{\nabla} \cdot \bar{\boldsymbol{u}} \mathbf{I} + \mu_c \bar{\mu} u_c L^{-1}\left(\bar{\nabla} \otimes \bar{\boldsymbol{u}} + \bar{\boldsymbol{u}} \otimes \bar{\nabla}\right) \tag{15}$$

By dividing both sides by $\mu_c u_c L^{-1}$, we obtain

$$\frac{\sigma_c}{\mu_c u_c L^{-1}} \bar{\boldsymbol{\sigma}} = \bar{\nabla} \cdot \bar{\boldsymbol{u}} \kappa \bar{\lambda} \mathbf{I} + \bar{\mu}\left(\bar{\nabla} \otimes \bar{\boldsymbol{u}} + \bar{\boldsymbol{u}} \otimes \bar{\nabla}\right) \tag{16}$$

where $\kappa$ equals $\lambda_c / \mu_c$. The scaling factor of the Cauchy stress tensor $\sigma_c$ equals $\mu_c u_c L^{-1}$.

The Navier-Cauchy equation for dynamic problems can be expressed as,

$$\nabla \otimes \left[ (\lambda + \mu) \nabla \cdot \boldsymbol{u} \right] + \nabla \cdot (\mu \nabla \otimes \boldsymbol{u}) = \rho \frac{\partial^2 \boldsymbol{u}}{\partial t^2} \tag{17}$$

Convert the parameters in **Eq. (17)** into their dimensionless form:

$$u_c L^{-2} \bar{\nabla} \otimes \left[ \left( \lambda_c \bar{\lambda} + \mu_c \bar{\mu} \right) \bar{\nabla} \cdot \bar{\boldsymbol{u}} \right] + u_c L^{-2} \bar{\nabla} \cdot \left( \mu_c \bar{\mu} \bar{\nabla} \otimes \bar{\boldsymbol{u}} \right) = u_c \rho_c \bar{\rho} \frac{\partial^2 \bar{\boldsymbol{u}}}{\partial t^2} \tag{18}$$

By dividing both sides by $u_c$, we obtain

$$\mu_c L^{-2} \left\{ \bar{\nabla} \otimes \left[ \left( \kappa \bar{\lambda} + \bar{\mu} \right) \bar{\nabla} \cdot \bar{\boldsymbol{u}} \right] + \bar{\nabla} \cdot \left( \bar{\mu} \bar{\nabla} \otimes \bar{\boldsymbol{u}} \right) \right\} = \rho_c \bar{\rho} \frac{\partial^2 \bar{\boldsymbol{u}}}{\partial t^2} \tag{19}$$

Based on **Eq. (19)**, it can be concluded that when the material is not subjected to its body force (such as gravity, electromagnetic force, etc.), $u_c$ does not affect the dimensionless Navier-Cauchy equation for dynamic problems. Therefore, in practical dynamic problems, $u_c$ can be assigned a constant value based on the specific situation.

The dimensionless PINN loss function $\mathscr{L}_{\text{PINN}}$ can be expressed as,

$$\mathscr{L}_{\text{PINN}} = \omega_{\text{OP}} \mathscr{L}_{\text{observation}} + \omega_{\text{PI}} \mathscr{L}_{\text{physical_information}} = \omega_{\text{OP}} \mathscr{L}_{\text{observation}} + \omega_{\text{PI}} \left( \mathscr{L}_{\text{boundary}} + \mathscr{L}_{\text{balanced}} + \mathscr{L}_{\text{constitutive}} \right) \tag{20}$$

with

$$\mathscr{L}_{\text{observation}} = \left| \sigma_{xx} - \sigma_{xx}^* \right|_{\text{MSE}} \tag{21}$$

$$\mathscr{L}_{\text{boundary}} = \left| u_x - u_x^* \right|_{\text{MSE}}^{\text{left}} + \left| u_x - u_x^* \right|_{\text{MSE}}^{\text{right}} + \left| u_y - u_y^* \right|_{\text{MSE}}^{\text{bottom}} \tag{22}$$

$$\mathscr{L}_{\text{balanced}} = \left| \mu_c \rho_c^{-1} L^{-2} \left( \frac{\partial \bar{\sigma}_{xx}}{\partial \bar{x}} + \frac{\partial \bar{\sigma}_{xy}}{\partial \bar{y}} \right) - \bar{\rho} \frac{\partial^2 \bar{u}_x}{\partial t^2} \right|_{\text{MSE}} + \left| \mu_c \rho_c^{-1} L^{-2} \left( \frac{\partial \bar{\sigma}_{xy}}{\partial \bar{x}} + \frac{\partial \bar{\sigma}_{yy}}{\partial \bar{y}} \right) - \bar{\rho} \frac{\partial^2 \bar{u}_y}{\partial t^2} \right|_{\text{MSE}} \tag{23}$$

$$\mathscr{L}_{\text{constitutive}} = \left| \bar{\sigma}_{xx} - \left( \kappa \bar{\lambda} + 2\bar{\mu} \right) \bar{\varepsilon}_{xx} - \kappa \bar{\lambda} \bar{\varepsilon}_{yy} \right|_{\text{MSE}} + \left| \bar{\sigma}_{yy} - \left( \kappa \bar{\lambda} + 2\bar{\mu} \right) \bar{\varepsilon}_{yy} - \kappa \bar{\lambda} \bar{\varepsilon}_{xx} \right|_{\text{MSE}} + \left| \bar{\sigma}_{xy} - 2\bar{\mu} \bar{\varepsilon}_{xy} \right|_{\text{MSE}} \tag{24}$$

where $\left| \cdot \right|_{\text{MSE}}$ is the mean squared error loss function, $\omega_{\text{OP}}$ and $\omega_{\text{PI}}$ are the weight of the observation points part and the physical information part of the loss function, respectively, $\bar{\boldsymbol{\varepsilon}} = \left( \bar{\nabla} \otimes \bar{\boldsymbol{u}} + \bar{\boldsymbol{u}} \otimes \bar{\nabla} \right) / 2$ is the dimensionless Cauchy strain tensor.

## 3. Results

### 3.1. Problem setup

In this section, the mechanical problems studied are focused on square TRMs with a length of 2mm. In the image of TRM microstructures, the elastic moduli of phase-1 and phase-2 are 20.6GPa and 206GPa, respectively; the Poisson's ratios of phase-1 and phase-2 are 0.2 and 0.4, respectively. The problem is two-dimensional. The boundary conditions and the position of dynamic displacement loads are shown in **Fig. 7a**. The magnitude of the dynamic displacement load changes with time as shown in **Fig. 7b**. The temporal evolution of the displacement magnitude at the location marked by the red dashed line is shown in **Fig. 7c**. The description of dynamic displacement loading in **Fig. 7** is only an example in datasets. The blue regions represent the TRM microstructures.

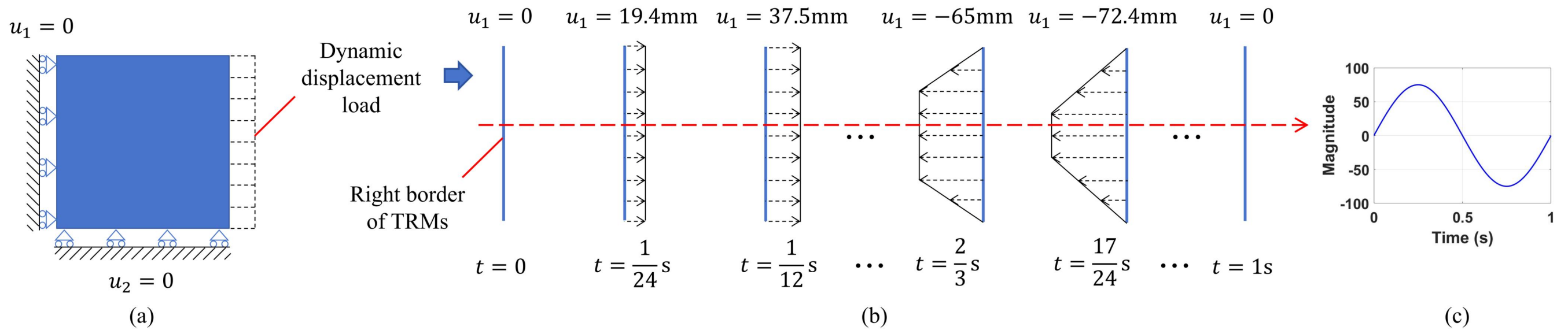

**Fig. 7.** Problem setup.

### 3.2. Results of STS-diffusion

The dataset consists of TRMs microstructure data generated by slicing a random field, dynamic displacement loading data, and stress data computed via the FEM, where the spatiotemporal resolution of the stress data is 24×64×64 [T, H, W]. The dataset sizes for $\sigma_{xx}$ are 2000, and they are split into training, validation, and test sets in an 8:1:1 ratio. To construct the pre-trained model, we randomly select 100 samples from $\sigma_{xx}$ dataset for model pre-training. The computational platform is configured as follows, OS: Windows Server 2019; CPU: Intel Xeon E5-2682 v4 @ 64x 3GHz; RAM: 32GB; GPU: NVIDIA GeForce RTX 3090 24GB.

**Fig. 8a**, **Fig. 8b**, and **Fig. 8c** respectively present the TRMs microstructure images, phase interfere information, and dynamic displacement loading information used for formal model testing. **Fig. 8d** and **Fig. 8e** show the spatiotemporal stress data generated by STS-diffusion and computed by FEM, respectively. **Fig. 8f** illustrates the error distribution between the STS-diffusion and FEM-computed spatiotemporal stress data. The mean error and the relative mean error (RME) are reported below the error distribution. Mean error and RME can be expressed as,

$$\text{Mean error} = \text{mean}\left(\text{Result}_{\text{STS-diffusion}} - \text{Result}_{\text{FEM}}\right) \tag{22}$$

$$\text{RME} = \frac{\text{mean}\left(\text{abs}\left(\text{abs}\left(\text{Result}_{\text{STS-diffusion}}\right) - \text{abs}\left(\text{Result}_{\text{FEM}}\right)\right)\right)}{\text{mean}\left(\text{abs}\left(\text{Result}_{\text{FEM}}\right)\right)} \tag{23}$$

where $\text{mean}(\cdot)$ and $\text{abs}(\cdot)$ denotes the average operator and absolute value operator, respectively.

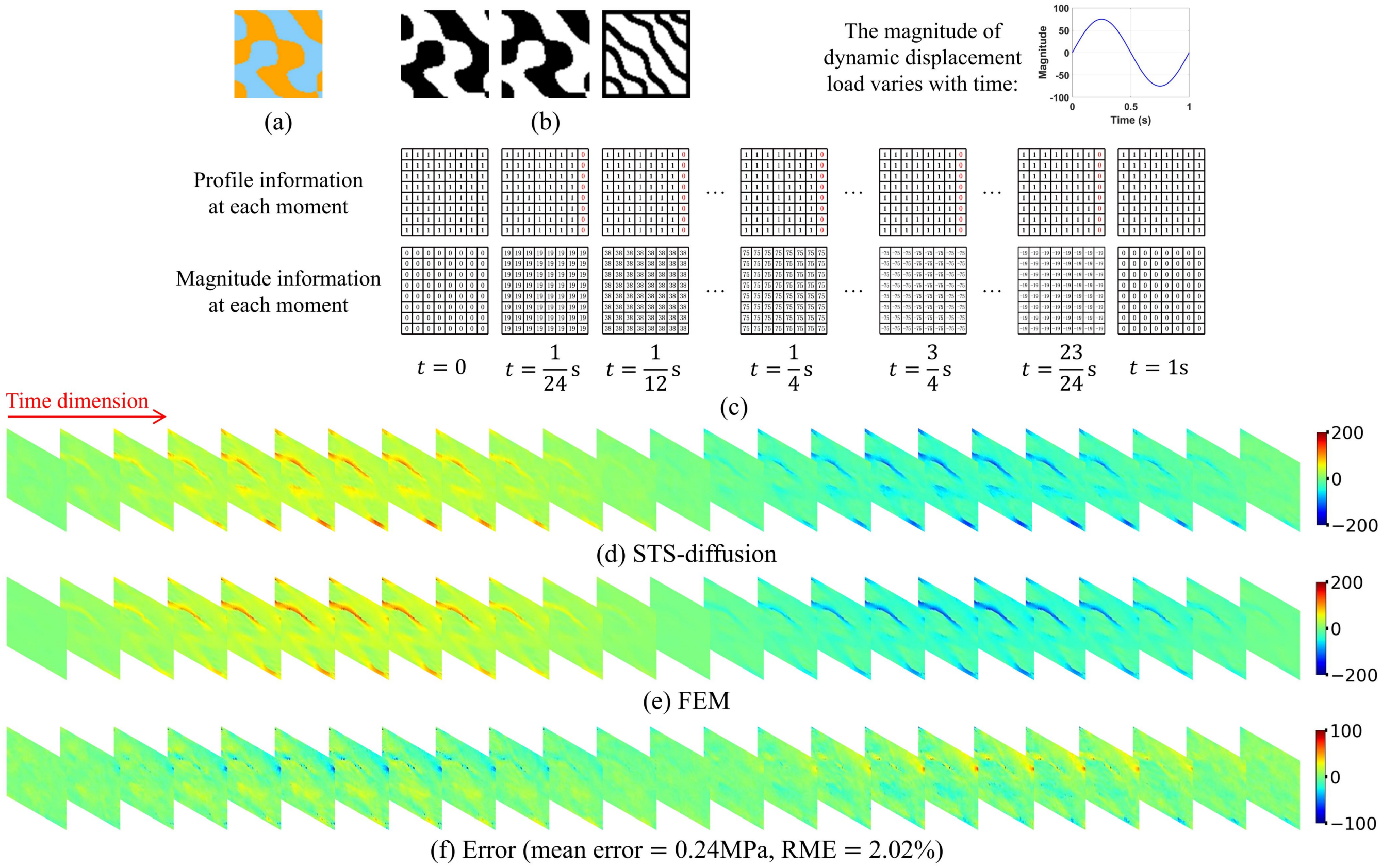

**Fig. 8.** Test results of spatiotemporal stress data. (a) TRM microstructure. (b) Corresponding phase interface information. (c) Dynamic displacement loading information. (d) STS-diffusion results. (e) FEM results. (f) Error.

To further investigate the impact of the attention position on the spatiotemporal stress data generation performance of STS-diffusion, we examined the effects of spatial attention and temporal attention at different insertion positions. A statistical analysis of the mean error between the generated spatiotemporal stress data and the FEM results across the entire test set, are summarized in Table 1. It is important to note that the training strategy for all models remains consistent, using the same pre-trained model. After hyperparameter tuning, the models are trained with a fixed learning rate of $1\times10^{-9}$. Both training and validation phases employ the mean squared error (MSE) as the loss function. The validation error is computed every 100 epochs, and the training process is terminated if the validation error fails to decrease for 10 consecutive evaluations. The model with the lowest validation error is selected for prediction and subsequent performance evaluation.

**Table 1.** Comparison of test set results between SR-MPINN and FEM under different attention positions and phase interface width information.

| **Attention positions** | **Train hours (h)** | **RME of $\sigma_{xx}$** |
|---|---|---|
| ① ② ③ ④ ⑤ ⑥ ⑦ | 11.67 | 2.43% |
| ② ③ ④ ⑤ ⑥ | 10.17 | 5.50% |
| ② ③ ⑤ ⑥ | 8.33 | 7.28% |
| ② ④ ⑥ | 6.02 | 8.61% |
| ③ ④ ⑤ | 6.64 | 8.26% |
| ④ | 5.21 | 7.98% |
| None | 4.86 | 9.35% |

The data presented in **Table 1** indicate that the position of attention mechanisms significantly impacts both model performance and training efficiency. In terms of relative mean error (RME) on the test set, the model incorporating attention at all positions (①–⑦) achieves the best performance (2.43%) but requires the longest training time (11.67 hours). This suggests that spatial and temporal attention effectively capture spatiotemporal features but come at a higher computational cost. When attention is applied only at positions ②–⑥, the RME increases to 5.50%, while training time is slightly reduced to 10.17 hours. The model without attention exhibits the highest RME (9.35%), confirming the necessity of the attention mechanism. It is recommended to prioritize the full-position attention configuration (①–⑦), as it reduces the RME by 55.8% compared to the ②–⑥ configuration, demonstrating a significant performance improvement. Although training time increases by 14.75% (1.5 hours), this overhead is acceptable given sufficient computational resources. For lightweight deployment, attention at positions ②–⑥ is a viable alternative, as it maintains a relatively efficient training time while outperforming other configurations. Conversely, applying attention only at position ④ provides suboptimal performance and is not recommended. These findings highlight the necessity of incorporating spatial attention to emphasize critical spatial features and temporal attention to capture dynamic variations, thereby enhancing the ability of model to represent complex spatiotemporal relationships effectively.

### 3.3. Results of ST-SRPINN

The observation data for ST-SRPINN is derived from the spatiotemporal stress data generated by STS-diffusion, comprising a total of 98,304 (24×64×64) observation points. The proposed ST-SRPINN architecture consists of: (1) two independent feedforward neural networks (FFNs) for approximating the displacement fields $u_x$ and $u_y$, each with 12 hidden layers and 64 neurons per layer; (2) three independent FFNs for approximating the stress fields $\sigma_{xx}$, $\sigma_{yy}$ and $\sigma_{xy}$, each with 16 hidden layers and 64 neurons per layer. Swish activation function is adopted for all networks. Both training and validation were performed using mean squared error. The number of training epochs is set to 1000.

After training, self-prediction is conducted on all observation points to generate spatiotemporal stress data with a resolution of 24×64×64. The generated results are then compared with FEM results, as illustrated in Fig. 9. The mean error and RME, computed with respect to the FEM results, are provided below the error data.

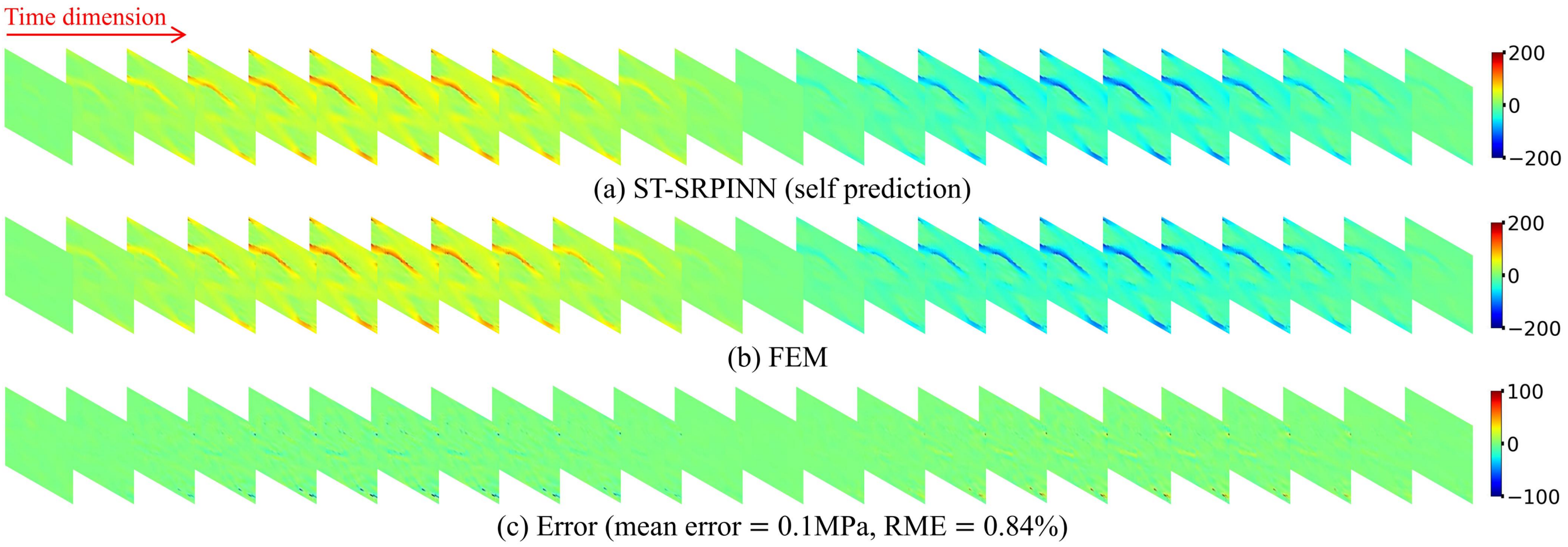

**Fig. 9.** The self-prediction results of ST-SRPINN.

It can be observed that after processing the generated results of STS-diffusion with ST-SRPINN, the self-predicted spatiotemporal stress data (with a resolution of $24\times64\times64$) exhibits a significantly lower mean error compared to traditional data-driven methods. This demonstrates that the incorporation of physics-informed constraints plays a crucial role in refining data-driven models, substantially enhancing both their accuracy and reliability. By leveraging physical constraints, this approach effectively mitigates the limitations of purely data-driven methods, further optimizing the generation performance. These results validate the importance and application potential of integrating physics-based priors in the generation of spatiotemporal data that inherently obeys physical laws.

The proposed method enables the spatiotemporal resolution of existing stress data to be upsampled by an arbitrary factor. To investigate the relationship between spatiotemporal resolution enhancement and the magnification factor, we conducted a series of experiments. Fig. 10 illustrates the results of spatiotemporal stress data upsampling with magnification factors of $1.25\times1\times1$, $1.25\times2\times2$, $1.25\times4\times4$, $2.5\times1\times1$, $2.5\times2\times2$, and $2.5\times4\times4$, yielding final resolutions of $30\times64\times64$, $30\times128\times128$, $30\times256\times256$, $60\times64\times64$, $60\times128\times128$, and $60\times256\times256$, respectively. The corresponding RME are provided beneath each super-resolved dataset.

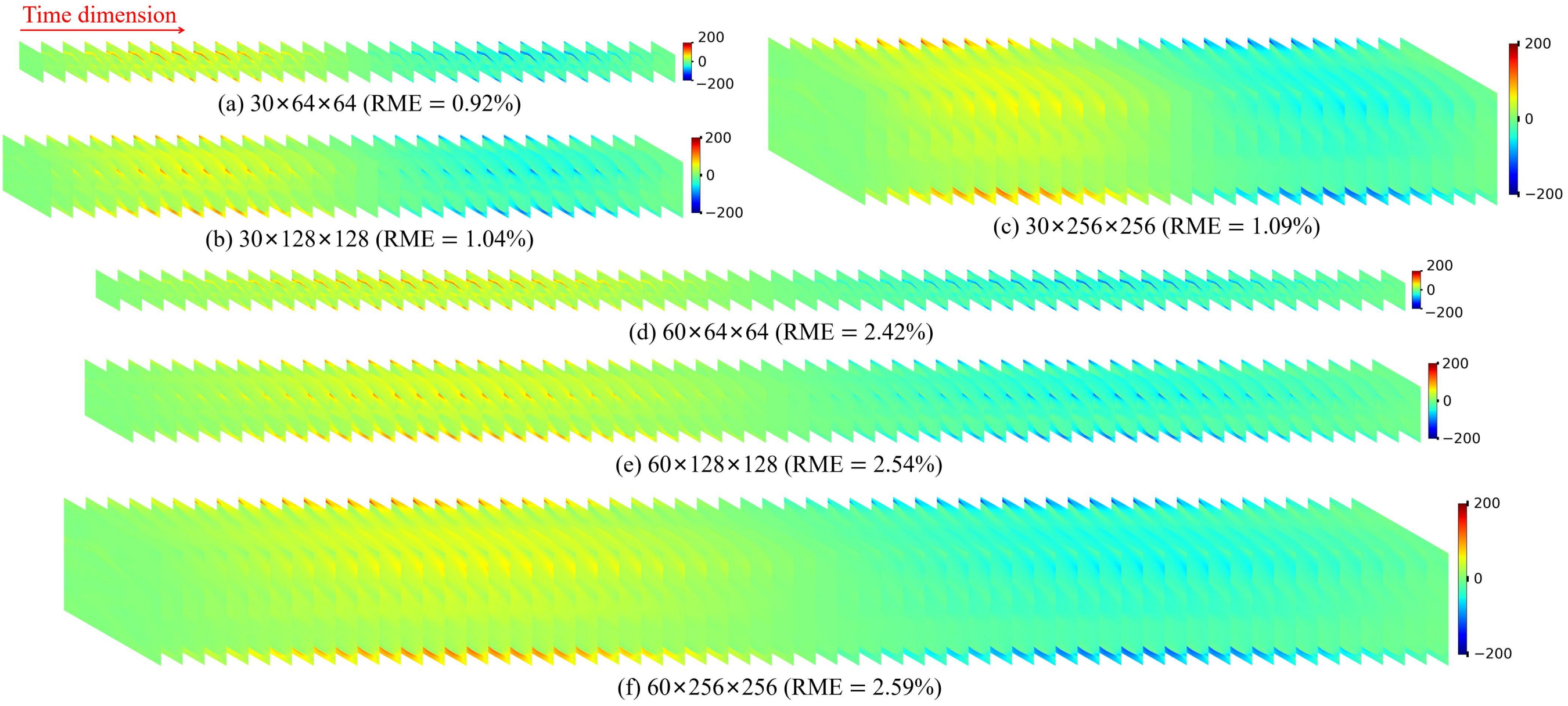

**Fig. 10.** ST-SRPINN results ( $\sigma_{xx}$ ) with different spatiotemporal resolutions. (a) 30×64×64. (b) 30×128×128. (c) 30×256×256. (d) 60×64×64. (e) 60×128×128. (f) 60×256×256.

The results indicate that as the spatiotemporal resolution increases, the RMS of the ST-SRPINN super-resolution results remains stable. This demonstrates the ability of the method to enhance resolution while preserving accuracy, validating its stability and robustness in handling spatiotemporal stress data.

Considering that different weights for the observation point and physical information components in the loss function may lead to different spatiotemporal super-resolution results (i.e., $\omega_{\mathrm{OP}}$ and $\omega_{\mathrm{PI}}$ in **Eq. (20)**). **Table 2** compares the errors of the ST-SRPINN and FEM spatiotemporal stress data results under different weight settings. The spatiotemporal stress data generated from the STS-diffusion model, where attention is applied at ①–⑦, is used for this comparison. It should be noted that the training strategy for all models remained consistent. After hyperparameter optimization, the model was trained with a learning rate of $1\times10^{-3}$ and the training epochs is set to 1000.

**Table 2.** Comparison of spatiotemporal stress results between SR-MPINN and FEM under different weighting conditions.

| **Weight ( $\omega_{\mathrm{OP}}:\omega_{\mathrm{PI}}$ )** | **Resolution** | **RME of $\sigma_{xx}$** |
|---|---|---|
| 5 : 1 | 30×64×64 | 1.26% |
| | 30×128×128 | 1.42% |
| | 30×256×256 | 1.48% |
| | 60×64×64 | 2.83% |
| | 60×128×128 | 2.99% |
| | 60×256×256 | 3.04% |
| 2 :1 | 30×64×64 | 1.39% |
| | 30×128×128 | 1.47% |
| | 30×256×256 | 1.55% |
| | 60×64×64 | 2.65% |
| | 60×128×128 | 2.73% |
| | 60×256×256 | 2.89% |
| 1 : 1 | 30×64×64 | 0.92% |
| | 30×128×128 | 1.04% |
| | 30×256×256 | 1.09% |
| | 60×64×64 | 2.42% |
| | 60×128×128 | 2.54% |
| | 60×256×256 | 2.59% |
| 1 : 2 | 30×64×64 | 0.82% |
| | 30×128×128 | 0.77% |
| | 30×256×256 | 0.83% |
| | 60×64×64 | 1.47% |
| | 60×128×128 | 1.67% |
| | 60×256×256 | 1.72% |
| 1 : 5 | 30×64×64 | 0.69% |
| | 30×128×128 | 0.45% |
| | 30×256×256 | 0.81% |
| | 60×64×64 | 0.79% |
| | 60×128×128 | 1.04% |
| | 60×256×256 | 1.07% |

The results indicate that the weight ratio between data-driven components (observation points) and physical information has a significant impact on the relative mean error (RME) of spatiotemporal stress data super-resolution. As the ratio of data-driven weight to physical information weight decreases from 5:1 to 1:5, the overall RME exhibits a downward trend. Notably, when the weight ratio reaches 1:5, the RMEs in the $\sigma_{xx}$ are significantly reduced, demonstrating strong error suppression capability. Furthermore, although RME generally increases with higher spatiotemporal resolution across different weight ratios, the error growth remains moderate without drastic fluctuations. This suggests that ST-SRPINN effectively integrates physical prior knowledge to enhance the accuracy of super-resolution results while ensuring system stability, preventing the accumulation of unstable errors caused by excessive reliance on data-driven components. Therefore, an appropriate adjustment of the data-driven and physical information weight ratio (e.g., $\omega_{\mathrm{OP}}:\omega_{\mathrm{PI}}=1:5$ ) is a key strategy for optimizing the super-resolution performance of ST-SRPINN.

# 4. Conclusions

This study develops a spatiotemporal stress analysis framework for two-phase random materials (TRMs), which can be applied to the dynamic analysis of materials with virtually any microstructural configuration. The specific conclusions are as follows:

A Generation method of spatiotemporal stress data based on diffusion model (STS-diffusion) is proposed. Considering the complexity of TRMs and the spatiotemporal nature of dynamic displacement loading, a condition embedding strategy is developed to effectively incorporate these factors. Additionally, to accommodate the high-dimensional characteristics of spatiotemporal stress data, we develop a tailored architecture for STU-net. To balance the trade-off between the generation accuracy and training efficiency of STS-diffusion, we investigate the position of attention mechanisms (including spatial attention and temporal attention) within STU-net. Our findings demonstrate that spatial and temporal attention effectively capture complex spatiotemporal dependencies.

A Spatiotemporal Super-Resolution Physics-Informed Operator of stress data (ST-SRPINN) is proposed, which is an unsupervised learning approach. This method requires only the spatiotemporal stress data generated by STS-diffusion as the training dataset and leverages physical constraints to upscale the resolution of spatiotemporal stress data to arbitrary magnification factors, including non-integer factors. We investigate the relationship between the upscaling factor and the accuracy of spatiotemporal stress data and observe that the super-resolution error converges, without a rapid increase as the magnification factor grows. Furthermore, we analyze the impact of the weight ratio between the data-driven components (observation points) and the physical constraints in the ST-SRPINN loss function on stress field enhancement and provide a recommended weight ratio ($\omega_{\mathrm{OP}}:\omega_{\mathrm{PI}}=1:5$).

In summary, the proposed framework integrates the construction of TRMs microstructures, the generation of spatiotemporal stress under dynamic loading, and the spatiotemporal super-resolution of stress data into a unified approach. Compared to previous stress generation methods, this framework not only addresses complex dynamic problems but also enables multiscale analysis of stress concentration regions at phase interferes. This enhancement significantly expands the engineering application potential of deep learning-based dynamic stress generation methods.

### Appendix A. Stochastic harmonic function (SHF) representation of random field

The stochastic harmonic function (SHF) allows for an accurate description of the desired power spectral density function using any number of superimposed terms.[31] The expression of SHF is as follows,

$$Y^{\mathrm{SHF}}(x_1,x_2)=\sum_{i=1}^{N_1}\sum_{j=1}^{N_2}A(\omega_{1,i},\omega_{2,j})\left[\cos\left(\omega_{1,i}x_1+\omega_{2,j}x_2+\varphi_{ij}^{\mathrm{I}}\right)+\cos\left(-\omega_{1,i}x_1+\omega_{2,j}x_2+\varphi_{ij}^{\mathrm{II}}\right)\right] \tag{A.1}$$

where $\varphi_{ij}^{\mathrm{I}}$ and $\varphi_{ij}^{\mathrm{II}}\left(i=1,2,\ldots,N_1;\,j=1,2,\ldots,N_2\right)$ denote the random phase independently identically distributed over $(0,2\pi]$. Assume the concerned wavenumber domain is $\Omega^{\mathrm{SHF}}=\left[\omega_1^{\mathrm{L}},\omega_1^{\mathrm{U}}\right)\times\left[\omega_2^{\mathrm{L}},\omega_2^{\mathrm{U}}\right)$, which was divided into non-overlapping sub-domains $\Omega_{i,j}^{\mathrm{SHF}}=\left[\omega_{1,i}^{\mathrm{L}},\omega_{1,i}^{\mathrm{U}}\right)\times\left[\omega_{2,j}^{\mathrm{L}},\omega_{2,j}^{\mathrm{U}}\right)$. The superscripts L and U represent the lower and upper bounds of the truncated wavenumber domain in the corresponding directions. $\omega_{1,i}$ and $\omega_{2,j}\left(i=1,2,\ldots,N_1;\,j=1,2,\ldots,N_2\right)$ are uniformly distributed in the intervals $\left[\omega_{1,i}^{\mathrm{L}},\omega_{1,i}^{\mathrm{U}}\right)$ and $\left[\omega_{2,j}^{\mathrm{L}},\omega_{2,j}^{\mathrm{U}}\right)$ respectively. The amplitude of each item can be expressed as

$$A(\omega_{1,i},\omega_{2,j})=\sqrt{4S(\omega_{1,i},\omega_{2,j})\left(\omega_{1,i}^{\mathrm{U}}-\omega_{1,i}^{\mathrm{L}}\right)\left(\omega_{2,j}^{\mathrm{U}}-\omega_{2,j}^{\mathrm{L}}\right)} \tag{A.2}$$

where $S(\omega_{1,i},\omega_{2,j})$ is the spectral density function of the Gaussian random field.

Assuming the autocorrelation function of the Gaussian stationary random field is exponential, its parameters can be determined by the two-point correlation function.[38] The autocorrelation function can be expressed as

$$R(\delta_1,\delta_2)=\exp\left[-\left(\frac{\delta_1}{c_1}\right)^2-\left(\frac{\delta_2}{c_2}\right)^2\right] \tag{A.3}$$

where $c_i\,(i=1,2)$ is the scaling factor associated with the scales of fluctuation or correlation lengths in the different directions.

According to the Wiener-Khinchin theorem, the power spectral density function can be expressed as

$$S(\omega_1,\omega_2)=\frac{c_1c_2}{4\pi}\exp\left[-\left(\frac{c_1\omega_1}{2}\right)^2-\left(\frac{c_2\omega_2}{2}\right)^2\right] \tag{A.4}$$

## Funding

This work was supported by the National Natural Science Foundation of China (Grant No.51538010).